\pdfoutput=1

\documentclass[10pt,twocolumn,letterpaper]{article}

\usepackage[pagenumbers]{cvpr} %

\usepackage{times}
\usepackage{epsfig}
\usepackage{graphicx}
\usepackage{amsmath}
\usepackage{amssymb}

\usepackage{listings}
\usepackage[table]{xcolor}
\usepackage{booktabs} %
\usepackage{colortbl}
\usepackage{glossaries}
\usepackage{algorithm}
\usepackage{multirow}
\usepackage[numbers]{natbib}
\usepackage[accsupp]{axessibility}  %
\usepackage[normalem]{ulem}
\usepackage[font=small,skip=4pt]{caption}

\usepackage{pifont}%
\usepackage{xspace}
\usepackage{framed}

\makeatletter
\let\inserttitle\@title
\makeatother

\usepackage{color, colortbl}
\definecolor{Gray}{gray}{0.9}
\definecolor{Goldenrod}{RGB}{230, 240, 255}%

\newcommand{\cmark}{\ding{51}\xspace}%
\newcommand{\xmarkg}{\textcolor{lightgray}{\ding{55}}\xspace}%

\def \ours          {SplitMask\xspace}

\usepackage[pagebackref,breaklinks,colorlinks]{hyperref}

\usepackage[capitalize]{cleveref}
\crefname{section}{Sec.}{Secs.}
\Crefname{section}{Section}{Sections}
\Crefname{table}{Table}{Tables}
\crefname{table}{Tab.}{Tabs.}

\def\cvprPaperID{637} %
\def\confName{CVPR}
\def\confYear{2022}

\newacronym{mim}{MIM}{Masked Image Modeling}
\newacronym{ours}{SplitMask}{SplitMask}

\begin{document}

\title{Are Large-scale Datasets Necessary for Self-Supervised Pre-training?}

\newcommand{\printfnsymbol}[1]{%
  \textsuperscript{\@fnsymbol{#1}}%
}

\author{%
\begin{minipage}{\linewidth}
\begin{center}
\scalebox{1.1}{\normalsize Alaaeldin El-Nouby\thanks{equal contribution}$^{\;,1,2}$ \hspace{0.35cm} Gautier Izacard$^{*,1,2}$  \hspace{0.35cm} Hugo Touvron$^{1,3}$ \hspace{0.35cm} Ivan Laptev$^{2}$}
\\[0.2cm]
\scalebox{1.1}{\normalsize Herv\'e J\'egou$^{1}$ \hspace{0.35cm} Edouard Grave$^{1}$} \\[0.5cm]
\scalebox{0.98}{\normalsize \textmd{$^1$Meta AI\hspace{0.6cm} $^2$Inria\hspace{0.6cm} $^3$Sorbonne University\hspace{0.6cm}}}
\\[1cm]
\end{center}
\end{minipage}
}

\maketitle

\begin{abstract}
Pre-training models on large scale datasets, like ImageNet, is a standard practice in computer vision. This paradigm is especially effective for tasks with small training sets, for which high-capacity models tend to overfit. 
In this work, we consider a self-supervised pre-training scenario that only leverages the target task data. 
We consider datasets, like Stanford Cars, Sketch or COCO, which are order(s) of magnitude smaller than Imagenet. 

Our study shows that denoising autoencoders, such as BEiT or a variant that we introduce in this paper, are more robust to the type and size of the pre-training data than popular self-supervised methods trained by comparing image embeddings. We obtain competitive performance compared to ImageNet pre-training on a variety of classification datasets, from different domains. 
On COCO, \textbf{when pre-training solely using COCO images}, the detection 
and instance segmentation 
performance surpasses the supervised ImageNet pre-training in a comparable setting. %

\end{abstract}

\section{Introduction}

\begin{figure}[t!]
    \centering
    \includegraphics[width=1.0\linewidth]{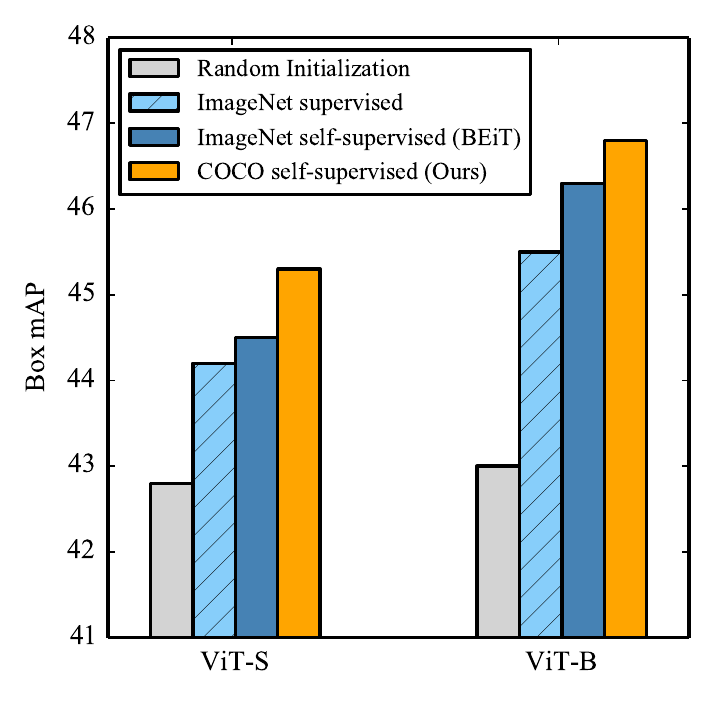}
    \caption{We demonstrate that self-supervised pre-training using denoising autoencoders like BEiT and our variant \ours are more robust to the type and/or size of pre-training data used. For example, the object detection performance of such models, when pre-trained only using COCO images and a Mask R-CNN pipeline, outperforms both supervised and BEiT self-supervised baselines pre-trained on ImageNet, as well as a randomly initialized baseline trained for a long schedule. %
    \label{fig:teaser}}
\end{figure}

Modern computer vision neural networks are heavily parametrized: they routinely have tens or hundreds of millions of parameters \cite{he2016deep,dosovitskiy2020image,liu2021swin,radosavovic2020designing}. %
This has been the key to their success for leveraging large-scale image collections such as ImageNet. 
However these high capacity models tend to overfit on small, or even medium sized datasets consisting of hundreds of thousands of images. 
This problem was pointed out by Oquab et al.~\cite{oquab2014learning} in 2014: 
\begin{center}
\scalebox{0.9}{
\begin{minipage}{1\linewidth}
\textcolor{black}{\it 
``Learning CNNs [...] amounts to estimating millions of parameters and requires a very large number of annotated image samples. This property currently prevents
application of CNNs to problems with limited training data.'' }
\end{minipage}}
\end{center}

The authors describe a learning setting~\cite{oquab2014learning,yosinski2014transferable} that is nowadays the dominant learning paradigm for data-starving problems: 

(1) pre-train a model on a large dataset like Imagenet~\cite{deng2009imagenet}, and in turn (2) finetune the weights of the models on the target task for which we have a limited amount of data. The second training stage typically adopts a shorter optimization procedure than the one employed when training from scratch (\ie, from randomly generated weights). 

This simple approach has led to impressive results, which are state-of-the-art in many tasks such as detection \cite{he2017mask, carion2020end}, segmentation \cite{chen2014semantic} and action recognition \cite{carreira2017quo}. Despite this success, we point out that it is difficult to disentangle the benefits offered by such a large-scale curated label dataset from the limitations of this pre-training paradigm. 
Putting aside the discussion on the collection effort (cost, requiring in-domain expertise, etc), we point out that  pre-training a model on a dataset and fine-tuning it on another can introduce two sort of discrepancies. 

First, this setting introduces a domain shift between the images used to pre-train the model and those targeted by the fine-tuning stage. Imagenet images may be sufficiently representative of natural images (despite the collecting bias). 
To date, most researchers consider that the benefit of having a large amount of images vastly compensates the domain discrepancy on benchmarks involving natural images, such as the fine-grained iNaturalist datasets~\cite{van2018inaturalist,Horn2019INaturalist} or even out-of-domain distributions such as sketches, painting or clipart.

The second question, discussed by Doersch \etal~\cite{doersch2020crosstransformers}, is the so-called \emph{supervision collapse}.
This phenomenon is inherent to pre-training with a fixed set of labels:
the network learns to focus on the mapping between images and the labels of the pre-training stage, but can discard information that is relevant to other downstream tasks.
In other terms, pre-training on large-scale classification datasets does not necessarily align with the goal of learning general-purpose features, %
as it uses only a subset of the available information controlled by the given dataset categorization bias \cite{ROSCH1973328}.

These limitations have motivated the development of self-supervised pre-training methods which learn directly from data, without relying on annotations. Most notably, the contrastive and joint embedding approaches  \cite{he2020momentum, swav, caron2021emerging, chen2020simple, grill2020bootstrap} can serve as effective pre-training strategies. While obtaining a strong performance on numerous tasks, such methods have a strong bias towards ImageNet data since the transformations have been hand-designed to perform well on the ImageNet benchmark. Some of the most effective transformations, like cropping, rely on the images being object centric \cite{purushwalkam2020demystifying}.
When applied on uncurated data, these methods degrade significantly and require larger datasets to obtain similar performance \cite{goyal2021self}.

This is in contrast with natural language processing, where nowadays, most applications use large models which were pre-trained on uncurated data.
In particular, the (masked) language modeling loss has been applied to transformer networks, leading to the BERT model~\citep{devlin2018bert},  which is now the foundation of most NLP models.
Inspired by this success, \citet{bao2021beit} have shown the potential of the \gls{mim} task to pre-train vision transformers. Such a model can be thought of as a denoising autoencoder \cite{vincent2008extracting} where the noise corresponds to the patch masking operation.
This technique has been successfully applied to ImageNet, but research questions remain: \smallskip

\noindent (1)\quad How much does this pre-training technique rely on the number of pre-training samples, and in particular, does it require millions of images to be useful?  %

\noindent (2)\quad Is this technique robust to different distributions of training images? In particular, is it an effective paradigm to learn with non object-centric or uncurated images? %

If the answer to both questions is positive, it will enable pre-training using a larger variety of datasets, including the training sets of many tasks that are smaller or belong to a different domain than ImageNet.

In this work, we make the following contributions:
\begin{itemize}
\item First, we demonstrate that denoising autoencoders 
are more sample efficient than joint embedding techniques, enabling pre-training without relying on large-scale datasets (e.g. ImageNet); 
\item Second, as a consequence of the better sample efficiency, we show on multiple datasets  that it is possible to pre-train directly on the target task data and obtain a competitive performance, even with datasets that are orders of magnitude smaller than ImageNet; %
\item Third, we demonstrate that denoising autoencoders can be successfully applied to non object-centric images such as COCO, achieving performance similar to the one obtained when pre-training with ImageNet, unlike joint embedding techniques which seem to suffer a drop in performance.

\end{itemize}

\begin{figure*}
    \begin{minipage}[t]{0.3\textwidth}
        \centering
        \includegraphics[ height=3.45cm,keepaspectratio]{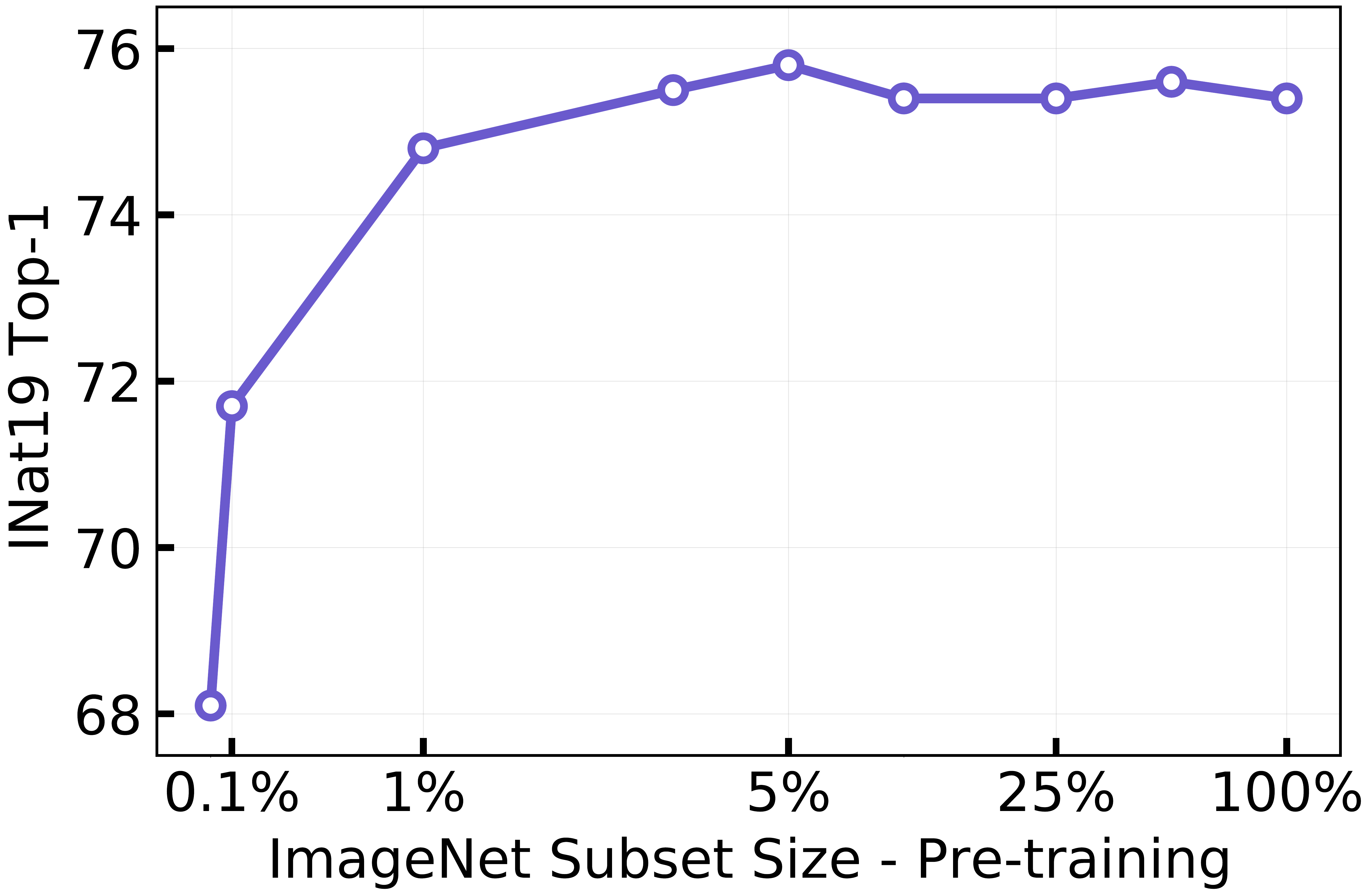}
    \end{minipage}%
    \hfill
    \begin{minipage}[b]{0.38\textwidth}
    \begin{minipage}{0.9\linewidth}
        \caption{Pre-training using different ImageNet subsets. Transfer performance does not improve beyond using a subset as small as 5\% when trained for the same number of iterations. \newline~ \newline~}
        \label{fig:imnet_subsets}  
    \end{minipage}
    
    \hspace{0.12\linewidth}
    \begin{minipage}[b]{0.88\linewidth}
        \caption{Varying the number of pre-training epochs for the 10\% subset. The performance first increases with longer training. Then we observe a plateau and a slight overfitting. }
        \label{fig:imnet_epochs}
    \end{minipage}
    \end{minipage}%
    \hfill \hfill
    \begin{minipage}[t]{0.3\textwidth}
        \centering
        \includegraphics[ height=3.45cm,keepaspectratio]{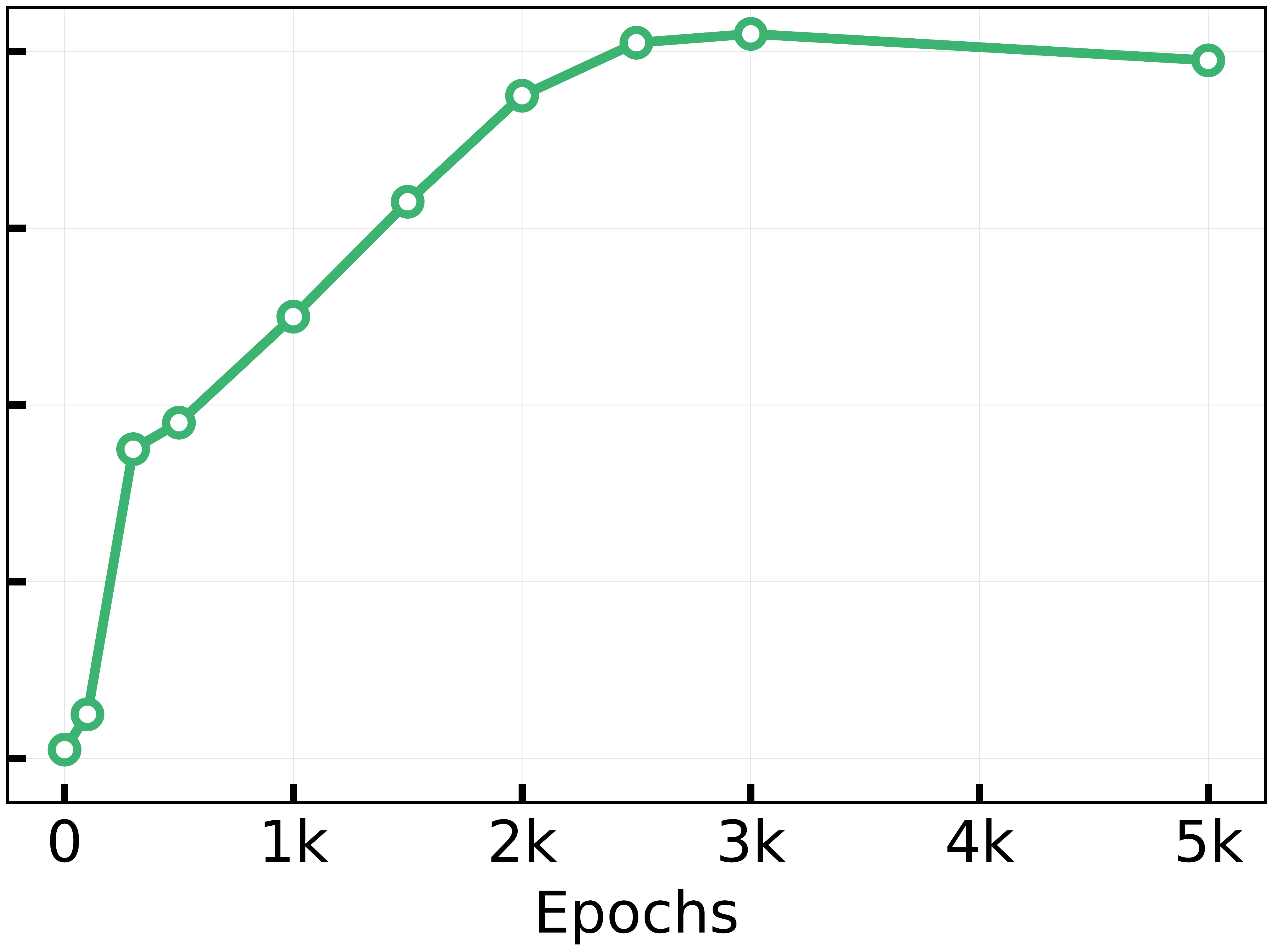}
    \end{minipage}%
\end{figure*}

\section{Related Work}
In this section, we briefly review some previous work on self-supervised learning, including autoencoders and instance discrimination methods.

\paragraph{Pre-training with autoencoders} has a long history in deep learning, where it was initially used as a greedy layer-wise method to improve optimization~\cite{hinton2006reducing,bengio2007greedy,ranzato2007efficient,vincent2010stacked, vincent2008extracting}.
In the context of unsupervised feature learning for image classification, different tasks related to denoising autoencoders have been considered, such as in-painting~\cite{pathak2016context}, colorization~\cite{zhang2016colorful} or de-shuffling of image patches~\cite{noroozi2016unsupervised}.
In natural language processing, denoising autoencoders have been applied by masking or randomly replacing some tokens of the input, and reconstructing the original sequence, leading to the BERT model~\cite{devlin2018bert}.
Similar methods have been proposed to pre-train sequence-to-sequence models, by considering additional kind of noises such as word shuffling or deleting~\cite{raffel2019exploring,lewis2019bart}.

There has been efforts to adopt such successful ideas in NLP  to computer vision, but with limited success. \citet{chen2020generative} proposed iGPT, a transformer-based autoregressive model that operates over image pixels, while \citet{atito2021sit} trained a ViT model on denoising of images where the noise is applied at pixel level.
More recently, \citet{bao2021beit} introduced the Masked Image Modeling loss  in computer vision, where image patches are masked, and the goal is to predict the discretized label of the missing patches corresponding to their visual words as defined by a pre-trained discrete VAE \cite{ramesh2021zero}.

\paragraph{Instance discrimination} is a set of self-supervised techniques which consider that each image corresponds to its own class~\cite{dosovitskiy2014discriminative,dosovitskiy2015discriminative}.
A set of data augmentations (or transformations) is then applied to each image to generate multiple examples for each class.
The global image representations are trained in a contrastive framework, typically using the InfoNCE loss~\cite{wu2018unsupervised}, to have high similarity for instances transformed from the same source image and low similarity with all other images.
As the performance of these methods depends on the number of negatives, it either requires large batches or memory banks to work well~\cite{wu2018unsupervised,he2020momentum,chen2020simple}.
It was later shown that when using a momentum encoder~\cite{he2020momentum}, simpler loss functions that did not directly discriminate against other images could be used~\cite{grill2020bootstrap,caron2021emerging, zbontar2021barlow, bardes2021vicreg, chen2021exploring}.
Finally, a related line of work is to use clustering techniques to pre-train deep neural networks~\cite{xie2016unsupervised,yang2016joint,caron2018deep,asano2019self,caron2020unsupervised}.

\paragraph{Transformer networks} were originally introduced in the context of machine translation, replacing recurrent neural networks by an attention-based mechanism~\cite{vaswani2017attention}.
Transformers were later applied to image recognition, by splitting images into patches, embedding these independently, and then processing the obtained representations as a sequence~\cite{dosovitskiy2020image}.
Initially, only vision transformers pre-trained on very large collections obtained good performance, but smaller models trained on ImageNet with heavy augmentation can also yield competitive tradeoffs~\cite{touvron2020deit}.

\paragraph{Pre-training data} is an important ingredient of self-supervised learning, and multiple works have studied its impact on the transfer performance of models.
While it is possible to learn high quality features from non-curated (eg. YFCC or IG) data using instance discrimination, this usually requires order of magnitude more data than ImageNet~\cite{caron2019unsupervised,goyal2021self}.
Similarly, one can perform supervised pre-training using weakly supervised data, such as using hashtags as labels, but this strategy also requires large amount of data to work well~\cite{joulin2016learning,mahajan2018exploring,dosovitskiy2020image}.
On the other hand, it was shown that for many natural language processing tasks, increasing the size of the pre-training dataset did not lead to strong improvement when using denoising autoencoders~\cite{raffel2019exploring}.
Finally, some work studied how much could be learned from a single pre-training image~\cite{asano2019critical} or from synthetic data~\cite{kataoka2020pre,krishna2021does}.

\section{Analysis}
In this section, we study the impact of the pre-training data on the performance of denoising autoencoder, and how they compare to those of joint embedding methods.
More precisely, we investigate how the number of images, and their nature, influence the quality of self-supervised models.
In this preliminary analysis, we consider the recent method BEiT and \ours,  our variant as detailed in Section \ref{sec:methods}, as representatives of denoising autoencoders, and DINO \cite{caron2021emerging} of a joint embedding method, respectively.

\begin{table}[t!]
  \centering
  \vspace{4mm}
  \caption{Analysis of different self-supervision methods transfer performance to the iNaturalist-2019 dataset when varying the size of the ImageNet subset used in the pre-training stage, in addition to using non object-centric dataset like COCO for pre-training. We observe that denoising autoencoders have a more robust behaviour w.r.t. pre-training data size or nature compared to joint embedding methods like DINO as well as supervised pre-training. }%
  \scalebox{0.8}{
    \begin{tabular}{l |  c  c  c | c}
      \toprule
       &  IMNet 1\% & IMNet 10\% & IMNet Full & COCO\\
      Method   & \textit{epochs: 30k} & \textit{epochs: 3k} & \textit{epochs: 300} &\textit{epochs: 3k}  \\ %
      \midrule
      Supervised  & 71.6 & 75.0 & 75.8 & \_ \\
      \midrule
      DINO \cite{caron2021emerging} & 70.1 & 73.1 & \textbf{78.4}  & 71.9 \\
      \rowcolor{Goldenrod}
      BEiT \cite{bao2021beit}  & 74.1 & 74.5 & 75.2 & 74.4 \\
      \rowcolor{Goldenrod}
      \ours  & \textbf{74.8} & \textbf{75.4} & 75.4 & \textbf{76.3} \\
      \bottomrule
  \end{tabular}}
  \label{tab:sample_efficiency}      
\end{table}

\subsection{Sample Efficiency}

\paragraph{Denoising autoencoders vs Supervised/DINO} First, we start by studying the impact of the pre-training dataset size, by varying the number of ImageNet examples we use to train models.
We consider subsets of ImageNet containing 10\% and 1\% of the total number of examples, and use the balanced (in terms of classes) subsets from \cite{assran2021semi}.
To decouple the effect of using smaller datasets and the effect of doing less training updates, we adapt the number of epochs to keep the number of iterations constant.
This means that we perform 3k and 30k epochs on ImageNet 10\% and 1\% respectively.
We report results in Table~\ref{tab:sample_efficiency}. Observe how pre-training with an autoencoder loss such as masked image modeling is robust to the reduction in dataset size.
In contrast, like for supervised pre-training, the performance of models pre-trained with DINO self-supervision degrades when training with smaller datasets.

\paragraph{Pre-training number of samples} We plot the iNaturalist-2019 transfer performance as a function of ImageNet subset size used during pre-training using \ours in Figure~\ref{fig:imnet_subsets}. We observe that the peak performance is achieved using only 5\% of the ImageNet samples and adding more samples does not provide additional boost, given the number of updates are kept constant. We also observe that using only a single image per class, which corresponds to the 0.1\% subset containing 1000 samples, leads to a non-trivial boost (+4 points) over training from scratch. This is a strong indication that denoising autoencoders are highly sample efficient unsupervised learning methods. 

\paragraph{Pre-training schedule length} Furthermore, we plot the transfer performance as a function of number of pre-training epochs in Figure~\ref{fig:imnet_epochs} using the 10\% ImageNet subset. It can be observed that training for long schedules of nearly 3k epochs, matching the total number of updates for that of full ImageNet with 300 epochs, is crucial to achieve such strong performance for smaller subsets. However, we observe slight overfitting for very long schedules. This problem is more predominant for pre-training using very small datasets like Stanford-Cars as illustrated in Figure~\ref{fig:cars_epochs}. 

\subsection{Learning using non object-centric images}
We now study the impact of changing the nature of the pre-training data. In particular we use images that are not object-centric, like in Imagenet. 
To this end, instead of pre-training using ImagetNet, we pre-train with images from the COCO dataset only.
As COCO contains roughly 118k images, this dataset is approximately equivalent in terms of size to the ImageNet 10\% subset.
Again, to disentangle the effect of training with a different number of iterations, we adapt the number of epochs: we use 3k epochs on COCO.

We report the results of this experiments in Table~\ref{tab:sample_efficiency}. 
When pre-trained on COCO, DINO drops significantly compared to full ImageNet pre-training (-8.3). Interestingly, the drop is higher than using 10\% ImageNet even though the numbers of samples is roughly the same. We hypothesis this is because COCO images are not biased to be object-centric, while this joint embedding method was designed and developed using ImageNet as benchmark. In contrast, BEiT's performance only decreases slightly while \ours attains +0.7 improvement over full ImageNet pre-training. This is an interesting property which makes such models prime candidates for learning  effectively from uncurated images in the wild.

\begin{table}[t]
    \centering
    \vspace{4mm}
    \caption{Ablation study on the effect of different tokenization methods.
      We compare the DALL-E tokenizer originally used in BEiT with patch level techniques: random projection, random patches and k-means clustering.
      We observe that the DALL-E tokenizer can be effectively replaced by simpler methods that do not require training on a large dataset.
      }
    \scalebox{0.9}{
      \begin{tabular}{lcccc}
        \toprule
        & DALL-E & Rand. Proj. & Rand. Patches & K-Means \\
        \midrule
        iNat19  & 75.2 & 75.2 & 75.3 & 75.0 \\
        \bottomrule
    \end{tabular}}
    \label{tab:tokenizers}
\end{table}

\subsection{Tokenizers}
The BEiT method, as proposed by \citet{bao2021beit}, relies on the discrete VAE tokenizer from DALL-E, which has been pretrained on a large weakly supervised dataset.
Since we want to study whether it is possible to pre-train models solely on small datasets, or non object-centric ones, we replace the DALL-E tokenizer by a simple alternative.
To this end, we consider different simple alternatives to discretize images at the patch level without any pre-training.
Each of these techniques is applied on each patch independently, making them relatively lightweight and more efficient than the original tokenizer considered in BEiT.

Given a vocabulary of size $V$, each element of the vocabulary is represented by a unit vector $\mathbf{e}_i \in \mathbb{R}^d$, where $i \in \{1, ..., V\}$ and
$d$ is the dimension of patches (in the case of 8x8 patches, $d=192$).
Then, to tokenize an image, we associate each patch to the element of the vocabulary which has the highest cosine similarity with the patch in the pixel space.
Hence, for a patch $\mathbf{x}$, its corresponding token $t$ is obtained as
\begin{equation}
t = \text{argmax}_{i \in \{1, ..., V\}} \mathbf{x}^\top \mathbf{e}_i.
\end{equation}

We now discuss three simple ways to obtain the elements of the vocabulary $\mathbf{e}_i$.
First, we can sample random vectors with uniform element-wise distribution, and call the corresponding tokenizer \emph{random projection}.
Second, we can sample $V$ random patches uniformly in the set of all patches of images from the training set, and refer to the tokenizer as \emph{random patches}.
Finally, we can perform k-means clustering on the patches of images from the training set, and use the centroids as elements of the vocabulary.
We refer to this last tokenizer, which was once widely employed in computer vision for bag-of-words representations, as \emph{k-means}.

We train a ViT-base model on the ImageNet dataset, using these three tokenizers, as well as the DALL-E tokenizer originally considered by BEiT.
We report results in Table~\ref{tab:tokenizers}.
We observe that replacing the DALL-E tokenizer by simpler choices does not lead to any significant degradation in accuracy.
This also provides a 26\% relative runtime improvement for base models over its counterpart using the DALL-E tokenizer on 16 GPUs with a batch size of 1024.

\begin{figure}[t!]
     \centering
     \includegraphics[trim={8.5cm 0.0cm 6.0cm 0.0cm}, width=1.02\linewidth]{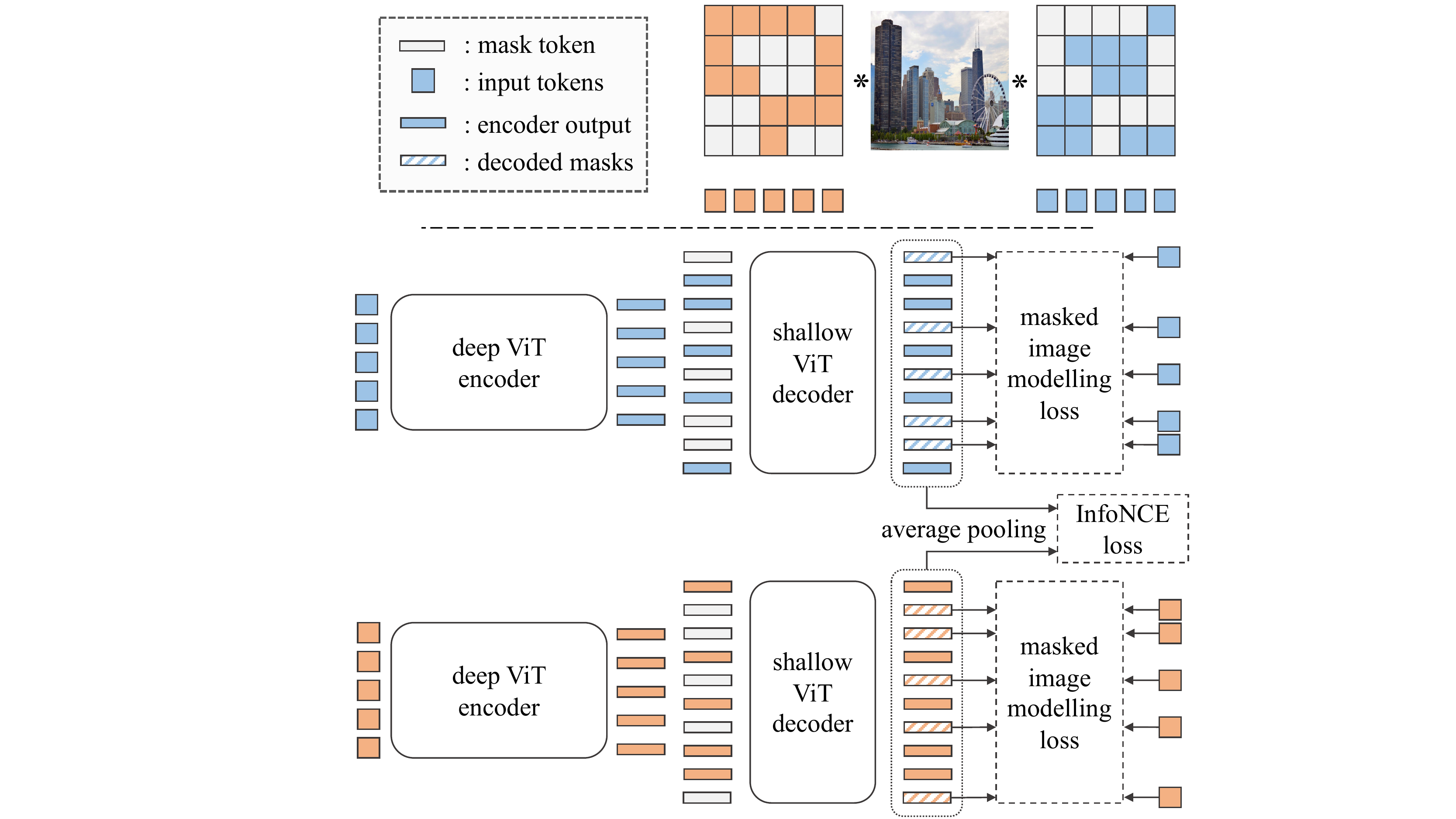}
    \caption{\ours consists of three steps. First, the input image patches are split into two disjoint subsets. Second, a shared deep ViT encoder processes each subset separately. The encoder outputs on each branch are augmented with a set of special mask tokens, representing the positions of the missing patches, and fed to a shallow ViT decoder. The decoder output corresponding to the mask tokens is used to solve a \gls{mim} task similar to BEiT. Finally, a global image descriptor is extracted from the decoder outputs of each branch by means of average pooling. The descriptors are trained to have high similarity using a contrastive loss (InfoNCE).
    \label{fig:overview}}
\end{figure}

\section{Methodology}
\label{sec:methods}
In this section, we introduce \ours, a variant of denoising autoencoders based on vision transformers. An overview of our method is illustrated in Figure~\ref{fig:overview}.

\subsection{\ours} 
Our approach is based on three steps, which we refer to as \emph{split}, \emph{inpaint} and \emph{match}.
As in standard vision transformers, an image is first broken down into patches of 16$\times$16 pixels.
Then, we \emph{split} the patches into two disjoint subsets $\mathcal{A}$ and $\mathcal{B}$, which are processed independently by our deep ViT encoder.
Next, using the patch representations of the subset $\mathcal{A}$ and a shallow decoder (e.g. 2 layers), we \emph{inpaint}\footnote{Inpainting in this context is implemented by solving a Masked Image Modeling task rather than the typical inpainting by reconstruction of pixels.}  the patches of the subset $\mathcal{B}$ , by solving a \gls{mim} task, and vice versa.
Finally, we obtain a global image descriptor by average pooling of the patch representations from the decoder output corresponding to  each branch. 

The feature aggregation is over both observed and hallucinated patches.
We try to \emph{match} the global descriptors of the image obtained from subset $\mathcal{A}$ to that obtained from subset $\mathcal{B}$.
In other words, we use the masking operation of the mask image modeling loss as a data augmentation for a contrastive learning loss similar to NPID or SimCLR. Note,  \ours does not add any significant computational cost over \gls{mim} methods like BEiT to produce this global contrastive training signal. 

\subsection{Encoder-Decoder Architecture}
\label{sec:enc_dec}

We now discuss in more details the architecture of the model that we use to implement the \ours pipeline described in the previous subsection.
Our method relies on an encoder-decoder architecture.
The encoder of our model is a standard vision transformer, with absolute positional embeddings.
In contrast to BEiT method, our encoder does not process representations of the \emph{masked} tokens, but only of the observed ones\footnote{Concurrent to our work, \citet{he2021masked} propose MAE. This is an encoder-decoder architecture where the encoder processing the observed patches only, similar to what we do in our \ours variant. }. 
Hence, an image is divided into patches, which are linearly embedded, and positional embeddings are added to these representations.
These representations are split into two subsets $\mathcal{A}$ and $\mathcal{B}$, which are processed independently by standard transformer layers.
Before feeding the output representations to the decoder, we insert mask embeddings that includes the position information of the missing patches in the sequences $\mathcal{A}$ and $\mathcal{B}$. 
Finally, using the decoded representations of the masked patches, we predict their corresponding visual words using a cross entropy loss function.

Thus, if an image contains $n$ patches, the encoder processes two sequences of size $n/2$, while the decoder processes two sequences of size $n$.
Since in practice we use decoder which is much more lightweight than standard vision transformers, the computational complexity of our models is similar to a standard ViT.
One advantage of our approach compared to BEiT is that at each iteration, the encoder processes all the patches of the image. The loss function is also computed over all the patches of the image, instead of only on a subset. Additional comparisons to BEiT are detailed in Sections~\ref{sec:splitmask_v_beit} and ~\ref{sec:enc_dec_app} of the appendix.

\subsection{Global Contrastive Loss}
In addition to the \gls{mim} loss, which is computed at the patch level, our approach also uses a contrastive loss at the image level.
To this end, we apply an average pooling operation over all the output representations of the decoder (including representations of the masked patches).
For each image, we obtain two representations $\mathbf{x}_{a}$ and $\mathbf{x}_b$, corresponding to the subsets $\mathcal{A}$ and $\mathcal{B}$ of observed patches.
We then apply the InfoNCE loss \cite{oord2018representation} over these representations:
\begin{equation}
\ell(\mathbf{x}_a) = \frac{\exp(\mathbf{x}_a^\top \mathbf{x}_b / \tau)}{\sum_{\mathbf{y} \in \{\mathbf{x}_b\} \cup \mathcal{N} } \exp(\mathbf{x}_a^\top \mathbf{y} / \tau)},
\end{equation}
where $\tau$ is a temperature hyper-parameter and $\mathcal{N}$ is a set of negatives, corresponding to the representations of the other images in the batch.
Following previous work~\cite{chen2020simple}, we symmetrize the contrastive loss, and apply it similarly on the representation $\mathbf{x}_b$ from the subset $\mathcal{B}$.
The motivation for adding this contrastive loss is to encourage the model to produce globally coherent features that are consistent across different choices of observed subsets without relying on any hand-designed transformations. Using our design of \ours, we attain such signal with almost no overhead.

\section{Experiments}
In this section, we perform empirical evaluations of denoising autoencoders, and the impact of the pre-training data on downstream task performance.
In particular, we study how well pre-training performs when only the target task data is used instead of relying on a large-scale dataset such as ImageNet.
We perform experiments on different tasks, such as classification, detection and instance segmentation. We consider datasets of varying size, including some significantly smaller than ImageNet.
We also compare our variant \ours method to BEiT, either pre-trained on target task data or ImageNet, in addition to the supervised pre-training baselines.
Finally, we perform an ablation study on our method to investigate the impact of its different components on finetuning and linear evaluation.

\subsection{Datasets}
We study the pre-training and finetuning of computer vision models on a variety of datasets, see Table~\ref{tab:datasets} for details.
For image classification, we consider the iNaturalist 2018 and 2019~\cite{van2018inaturalist}, Stanford Cars~\cite{Cars2013} and Food101~\cite{bossard14} datasets, which all contain fine-grained categories.
We also consider three subsets from the DomainNet dataset~\cite{peng2019moment}, \emph{clipart}, \emph{painting} and \emph{sketch}, which are not natural images and hence from different domains than ImageNet.
For object detection and instance segmentation, we use the COCO dataset~\cite{lin2014microsoft}.
Finally, we also use the ADE20k dataset~\cite{zhou2017scene} for semantic segmentation.
The training set sizes of these different datasets vary from 8k to 437k images, thus all being significantly smaller than ImageNet, some more than two order of magnitude smaller.
This allows to investigate under different data regimes how feasible it is to pre-train directly on the target task data, alleviating the need for a large scale curated dataset as ImageNet.

As previously mentioned, we want to perform a constant number of updates during pre-training, and we thus adapt the number of epochs when training on target task data to match the number of updates corresponding to 300 epochs on ImageNet. For smaller classification datasets, we limit the number of pre-training epochs to 5000 since we observed pre-training for longer generally does not result in further improvement in terms of downstream performance. For very small datasets, like Stanford Cars, we observed an overfitting behaviour with training for very long schedules (e.g. more than 5k epochs, see Figure~\ref{fig:cars_epochs}).
Note that the adjusted number of pre-training epochs is provided in Table~\ref{tab:datasets}.

\begin{table}[t!]
        \centering
        \vspace{4mm}
        \caption{Data size, number of classes and number of pre-training epochs details for all datasets used for pre-training.}
        \scalebox{0.8}{
        \begin{tabular}{l | r r r | r}
        \toprule
        Dataset & \#Train & \#Test & \#Classes  & Epochs \\
        \midrule
         ImageNet \cite{deng2009imagenet} & 1,281,167 & 50,000 & 1000 & 300 \\
         \midrule
         iNaturalist 2018 \cite{van2018inaturalist} & 437,513 & 24,426 & 8,142 & 800 \\
         iNaturalist 2019 \cite{Horn2019INaturalist} & 265,240 & 3,003 & 1,010 & 1,400\\
         Food 101 \cite{bossard14} & 75,750 & 25,250 & 101 & 5,000\\
         \midrule
         Stanford Cars \cite{Cars2013} & 8,144 & 8,041 & 196 & 5,000\\
         Clipart \cite{peng2019moment} &  34,019 & 14,818 & 345 & 5,000 \\
         Painting \cite{peng2019moment} &  52,867 & 22,892 & 345  & 5,000 \\
         Sketch \cite{peng2019moment} &  49,115 & 21,271 & 345  & 5,000 \\
         \midrule
          ADE20k~\cite{zhou2017scene} & 20,210 &  2,000 & 150 & 21,000 \\
          COCO \cite{lin2014microsoft} & 118,287 & 5,000 & 80 & 3,000 \\
         \bottomrule
        \end{tabular}}
        \label{tab:datasets}

\end{table}

\begin{table*}[t!]
    \centering
    \caption{COCO detection and instance segmentation performance, using a Mask R-CNN pipeline, for models with different pre-training recipes. We see that BEiT and \ours pre-training using COCO outperform supervised ImageNet pre-training of DeiT as well as self-supervised ImageNet pre-training using BEiT.  $\dagger$: Method uses a longer 6x schedule instead of the default 3x following \citet{he2018rethinking}. %
    }
    \scalebox{0.94}{
\begin{tabular}{l| c@{\ \ \ }|c | c c |c@{\ \ \ }c@{\ \ }c|c@{\ \ }c@{\ \ } c}
        \toprule
        Method & Backbone &  \multicolumn{3}{c |}{Pre-training}  &$\text{AP}^{b}$ & $\text{AP}^{b}_{50}$ & $\text{AP}^{b}_{75}$ & $\text{AP}^{m}$ & $\text{AP}^{m}_{50}$ & $\text{AP}^{m}_{75}$\\ 
        \cline{3-5}
         & & Supervised & IMNet & COCO & & & & & & \\
        \midrule 
        Random Initialization &  \multirow{6}{*}{ViT-S} & \xmarkg & \xmarkg & \xmarkg  & 38.3 & 60.1 & 41.4 & 35.6 & 57.1 & 37.7 \\
        Random Initialization$\dagger$ &  &\xmarkg & \xmarkg & \xmarkg & 42.8 & 64.5 & 45.6 & 39.1 & 61.5 & 41.7  \\
        DeiT \cite{touvron2020deit} &  &\cmark & \cmark & \xmarkg & 44.2 & 66.6 & 47.9 & 40.1 & 63.2 & 42.7\\
        BEiT \cite{bao2021beit} &  &\xmarkg & \cmark & \xmarkg & 44.5 & 66.2 & 48.8 & 40.3 & 63.2 & 43.1 \\
        DINO \cite{caron2021emerging} &  &\xmarkg & \xmarkg & \cmark &  43.7 & 65.5 & 47.7 & 39.6 & 62.3 & 42.3 \\
        \rowcolor{Goldenrod}
        BEiT &  &\xmarkg & \xmarkg & \cmark & 44.7 & 66.3 & 48.8 & 40.2  & 63.1 & 43.2 \\
        \rowcolor{Goldenrod}
        \ours & &\xmarkg & \xmarkg & \cmark & \textbf{45.3} & \textbf{66.9} & \textbf{49.4} & \textbf{40.6} & \textbf{63.6} & \textbf{43.5} \\
        \midrule
        Random Initialization &  \multirow{6}{*}{ViT-B} &\xmarkg & \xmarkg & \xmarkg & 40.7 & 62.7 & 44.2 & 37.1 & 59.1 & 39.4  \\
        Random Initialization$\dagger$ &   &\xmarkg & \xmarkg & \xmarkg & 43.0 & 64.2 & 46.9 & 38.8 & 61.3 & 41.6 \\
        DeiT \cite{touvron2020deit} &  &\cmark & \cmark & \xmarkg & 45.5 & \textbf{67.9} & 49.2 & 41.0 & 64.6 & 43.8\\
        BEiT \cite{bao2021beit} &  &\xmarkg & \cmark & \xmarkg & 46.3 & 67.6 & 50.6 & 41.6 & 64.5 & 44.9 \\
        DINO \cite{caron2021emerging} &  &\xmarkg & \xmarkg & \cmark & 43.1 & 64.4 & 46.9 & 38.9 & 61.4 & 41.4  \\
        \rowcolor{Goldenrod}
        BEiT &  &\xmarkg & \xmarkg & \cmark & 46.7 & 67.7 & 51.2 & 41.8 & 65.0 & 44.6  \\
        \rowcolor{Goldenrod}
        \ours & &\xmarkg & \xmarkg & \cmark  & \textbf{46.8} & \textbf{67.9} & \textbf{51.5} & \textbf{42.1} & \textbf{65.3} & \textbf{45.1} \\
        \bottomrule     
    \end{tabular}}
    \label{tab:coco}

\end{table*}

\subsection{Dense Prediction}

\subsubsection{Object detection and Instance Segmentation}
First, we evaluate our approach on the COCO object detection and instance segmentation dataset using the Mask R-CNN pipeline \cite{he2017mask} and report our results in Table~\ref{tab:coco}.
We compare models pre-trained on the COCO dataset alone with their equivalent counterparts that were pre-trained on ImageNet, either in a supervised or self-supervised fashion.
First, we observe that BEiT models which were pre-trained on the COCO dataset alone obtain better downstream task performance than the same models pre-trained on ImageNet.
For example, when using a ViT-base backbone, pre-training on COCO instead of ImageNet leads to a boost of +0.4 in box AP.

Additionally, we observe that a similar pre-training of DINO using COCO images provides a relatively weak performance, only outperforming random initialization. This indicates that strong pre-training on COCO is a unique property of denoising autoencoders  and it does not extend to other self-supervised learning methods.

Finally, we observe that \ours leads to a consistent improvement compared to the BEiT baseline, such as +0.6 box AP when using a ViT-small and +0.3 mask AP for ViT-base backbones. All put together, in a comparable setting, we obtain a +1.1 box AP increase while not using ImageNet. Since COCO contains one order of magnitude less images than ImageNet, this suggests that large scale datasets are not necessary for pre-training.

\subsubsection{Semantic Segmentation}

For semantic segmentation, we compare our denoising autoencoder models, pre-trained solely using ADE20k images, to their counterparts pre-trained on ImageNet. The results are reported in Table~\ref{tab:ade}. All models use an UperNet pipeline~\cite{xiao2018unified}.
We observe that denoising autoencoders can provide a very competitive performance on such a challenging task even when pre-trained using a relatively small sample size of 20k images. The performance matches that of BEiT self-supervised pre-training using ImageNet and only marginally lower than supervised ImageNet pre-training. 

We have found that adapting the random cropping strategy is a crucial implementation detail that helps improve the denoising autoencoders pre-training performance on such dataset.  In particular, we reduce the maximal size of the crop from 100\% to 25\% of the raw image size.

\begin{table}[t!]
    \centering
    \vspace{4mm}
    \caption{
    Semantic segmentation performance for different pre-trained models on ADE20k using an UperNet pipeline \cite{xiao2018unified}. All models reported use a ViT-B architecture.
    In spite of the small size of the ADE20k dataset, performance of our models provides a performance competitive to those pre-trained using ImageNet.
    }
    \scalebox{0.9}{
\begin{tabular}{l| c|c c|c}
        \toprule
        Method &\multicolumn{3}{c |}{Pre-training}  & mIoU \\ %
        \cline{2-4}
         &  Supervised & IMNet & ADE20k &   \\
         \midrule
        Random Init. &  \xmarkg & \xmarkg & \xmarkg & 25.4\\
        DeiT \cite{touvron2020deit} & \cmark & \xmarkg & \xmarkg & 46.1 \\
        BEiT \cite{bao2021beit} & \xmarkg & \cmark & \xmarkg & 45.6\\
        \rowcolor{Goldenrod}
        BEiT & \xmarkg & \xmarkg & \cmark & 45.6   \\
        \rowcolor{Goldenrod}
        \ours &\xmarkg & \xmarkg & \cmark & 45.7  \\
        \bottomrule     
    \end{tabular}}
    \label{tab:ade}

\end{table}

\begin{table*}[t!]
  \centering
  \caption{Comparison between finetuning performance on the target datasets of different sizes and domains when pre-trained using the target datasets themselves,  ImageNet pre-training (both supervised and self-supervised), and training from scratch.
  Both denoising autoencoders (BEiT and \ours) obtain competitive performance when solely using the target data. $\mp$: \citet{liu2021efficient} use a different pre-training setup and backbones.}
        \scalebox{0.9}{
        \begin{tabular}{l| c | c | c c | c c c | c c c c }
            \toprule
            Method & Backbone & Supervised & \multicolumn{2}{c|}{Data Used} & iNat-18 & iNat-19 & Food 101 & Cars  & Clipart & Painting & Sketch \\ %
            \cline{4-5}
            \cline{6-12}
            & & pre-training & IMNet & Target & 437k & 265k & 75k & 8k & 34k & 52k & 49k \\
             \midrule
             \multirow{2}{*}{\citet{liu2021efficient}}$\mp$ & CVT-13 & \xmarkg &  \xmarkg & \cmark & \_ & \_ & \_ & \_ & 60.6 & 55.2 & 57.6   \\
              & ResNet-50 & \xmarkg &  \xmarkg & \cmark & \_ & \_ & \_ & \_ & 63.9 & 53.5 & 59.6  \\
             \midrule
             Random Init. & \multirow{5}{*}{ViT-S} &  \xmarkg & \xmarkg & \cmark & 59.6 & 67.5 & 84.7 & 35.3  & 41.0 & 38.4 & 37.2 \\
             DeiT \cite{touvron2020deit} &  & \cmark & \cmark & \cmark & \underline{69.9} & 75.8 & \textbf{91.5} & 92.2 & \textbf{79.6} & \textbf{74.2} & \textbf{72.5} \\ %
              BEiT  \cite{bao2021beit} & &  \xmarkg &   \cmark & \cmark  & 68.1 & 75.2 & 90.5 & 92.4  & 75.3 & 68.7 & 68.5 \\
            \rowcolor{Goldenrod}
             BEiT & &  \xmarkg &  \xmarkg & \cmark & 68.8 & \underline{76.1} & 90.7 & \underline{92.7}  & \_ & 69.0 & \_ \\ %
             \rowcolor{Goldenrod}
             \ours & &  \xmarkg &  \xmarkg & \cmark & \textbf{70.1} & \textbf{76.3} & \textbf{91.5} & \textbf{92.8}  & \underline{78.3} & \underline{69.2} & \underline{70.7} \\ %
             \midrule
              Random Init. &  \multirow{5}{*}{ViT-B} & \xmarkg &  \xmarkg & \cmark & 59.6 & 68.1 & 83.3 & 36.9  & 41.9 & 37.6 & 34.9 \\
              DeiT \cite{touvron2020deit} & & \cmark &  \cmark & \cmark & \underline{73.2} & 77.7 & 
             \textbf{91.9} & 92.1  & \textbf{80.0} & \textbf{73.8} & \textbf{72.6} \\
              BEiT \cite{bao2021beit} & & \xmarkg &  \cmark & \cmark & 71.6 & 78.6 & 91.0 & \textbf{93.9}   & 78.0 & 71.5 & 71.4 \\

            \rowcolor{Goldenrod}
             BEiT & & \xmarkg &  \xmarkg & \cmark & 72.4 & \underline{79.3} & \underline{91.7} & 92.7  & \_ & 70.7 & \_ \\
             \rowcolor{Goldenrod}
             \ours & & \xmarkg &  \xmarkg & \cmark & \textbf{74.6} &\textbf{80.4} & 91.2 & \underline{93.1} & \underline{79.3} & \underline{72.0} & \underline{72.1} \\
            \bottomrule
        \end{tabular}}
        \label{tab:classification}
\end{table*}

\begin{table}[t!]
  \centering
  \vspace{4mm}
  \caption{Finetuning performance on ImageNet. Here, epochs refer to the number of pre-training epochs on ImageNet.}
  \scalebox{0.9}{
    \begin{tabular}{l | c c | c}
      \toprule
      Method & Backbone & Epochs & Top-1  \\
      \midrule
      MocoV3 \cite{chen2021empirical} & \multirow{4}{*}{ViT-S} & 300 & 81.4  \\
      DINO \cite{caron2021emerging} & & 300 & \textbf{81.5}  \\
      BEiT \cite{bao2021beit} & & 300 & 81.3 \\
      \rowcolor{Goldenrod}
      \ours & & 300 & \textbf{81.5} \\
      \midrule
      MocoV3 \cite{chen2021empirical} & \multirow{4}{*}{~ \newline ViT-B} & 300 & 83.2 \\
      DINO \cite{caron2021emerging} & & 400 &\textbf{83.6}  \\
      BEiT \cite{bao2021beit} & & 300 & 82.8 \\
      BEiT  \cite{bao2021beit} & & 800 & 83.2 \\
      \rowcolor{Goldenrod}
      \ours & & 300 & \textbf{83.6} \\
      \bottomrule
  \end{tabular}}
  \label{tab:ssl_compare}
\end{table}

\subsection{Image Classification}

We perform empirical evaluation on a number classification datasets and report our results in Table~\ref{tab:classification}. Overall, we find that BEiT or \ours pre-training, using solely the target datasets images, consistently obtains either the strongest or, at worst, the second strongest performance when compared to different options of self-supervised and supervised pre-training using ImageNet as well as training from scratch \cite{liu2021efficient}. 

\paragraph{BEiT pre-training: ImageNet vs Target}
First, we compare ImageNet pre-training to the target data pre-training with BEiT and observe that for many cases, pre-training on the target data alone leads to better results. %
This is true for the ViT-small backbone across all the datasets including Stanford cars (+1.1\% acc), which consists of only 8k images.
When using a ViT-base backbone, pre-training on the target task data outperforms BEiT self-supervised ImageNet pre-training for datasets as small as Food101 (+0.7 acc), which is more than 10x smaller than ImageNet.
Second, we observe that \ours leads to further improvement in performances for multiple datasets:
for example, on the iNaturalist 2018 dataset, we see +3.0 in accuracy with a ViT-base model.

\paragraph{Supervised ImageNet pre-training}
As it was already observed in previous work~\cite{chen2020simple,chen2021empirical,caron2021emerging}, we also see in many cases that self-supervised training outperforms supervised pre-training on ImageNet.
For example, on the iNaturalist datasets, training with the target task data alone (including a pre-training step) gives better results than pre-training on ImageNet with labels:
with a ViT-base model and the \ours method, we see an improvement of +2.7\% in top-1 accuracy. %
As for the \emph{clipart}, \emph{painting} and \emph{sketch} datasets, we see that \ours provides a competitive performance, outperforming an ImageNet pre-trained BEiT across all datasets for ViT-S.
However, for the aforementioned  datasets, supervised pre-training achieves the best performance for both ViT-S and ViT-B.

 We note that when pre-training using the \emph{clipart} and \emph{sketch} datasets with the BEiT method, we experienced numerical instability that prevented the model from converging with long schedules (e.g. 5000 epochs). However, the instability problem was not observed for \ours models. Nevertheless, more investigation might be needed to fully understand how to optimize pre-training of such models.

\subsection{Pre-training using ImageNet}

In addition to our main study concerning the robustness of denoising autoencoders w.r.t the size and type of pre-training data, we study \ours in the more commonly used setting of pre-training and finetuning using ImageNet.

In Table~\ref{tab:ssl_compare} we show the performance of our \ours method using the ViT-S and ViT-B backbones and 300 epochs pre-training compared to other recent transformer-based self-supervised learning methods. It can be observed that \ours provides a strong performance, outperforming both BEiT and MocoV3 for all backbones. Additionally, \ours achieves a performance on par with DINO while being significantly cheaper and simpler to train. Note that while \ours and BEiT attain a strong finetuning performance, denoising autoencoding methods typically fall behind in terms of linear probing compared to instance discrimination methods like DINO.

\subsection{Implementation Details}

\paragraph{Tokenizers.}
Similarly to the tokenizer used in~\citep{bao2021beit}, all tokenizers presented in Table~\ref{tab:tokenizers} have a vocabulary of size 8192.
For the random tokenizer, we sample 8192 vectors with uniform component-wise distribution.
For the random patches tokenizer we sample 8192 patches from different images.
For the K-means tokenizer, the 8192 elements of the vocabulary are obtained by applying the K-means algorithm to 3 millions patches sampled from the dataset.

\paragraph{Pre-training.} We use the original ViT formulation as proposed by \citet{dosovitskiy2020image} and we follow the pre-training hyperparameters of \citet{bao2021beit}. All baselines reported use the same backbone implementation and trained in similar settings. For \ours, by default, we use random block masking \cite{bao2021beit} of 50\% masking ratio to obtain a mask and its complement to extract the two subsets. The maximum and minimum number of patches per block is 75 and 16 respectively. We use the standard random cropping and horizontal flipping as data augmentations. We use 2 transformer layers for the decoder with embedding dimension matching that of the encoder. 

However, for the smallest datasets (i.e. Stanford-Cars, ClipArt, Sketch and Paintings), we found that stronger data augmentation and more aggressive masking prevents early overfitting. In particular, we use a uniform masking of 75\% (like in the work by He et al.~\cite{he2021masked}), as well as using random greyscale, solarization, Gaussian blur and color jittering as additional forms of data augmentation. 

The BEiT baselines pre-trained on ImageNet and reported in Table~\ref{tab:coco} and~\ref{tab:classification} use the DALL-E tokenizer. 
Other BEiT and \ours models have been pre-trained using our random projection tokenizer.
For the InfoNCE loss we use $\tau=0.2$ following \citet{chen2021empirical}.

\paragraph{Object detection and Instance segmentation.} We use the Mask R-CNN detection method~\cite{he2017mask} with ViT backbone as our detection method. In order to obtain features compatible with the Feature Pyramid Network (FPN) design \cite{lin2017feature}, we use max pooling and transposed convolution operations similar to \citet{el2021xcit}. To accommodate for the variable resolution we replace the absolute positional encoding for our models and the baselines with sinusoidal positional encoding \cite{vaswani2017attention}. All models are trained using the 3x schedule (36 epochs) unless mentioned otherwise. We use the training hyper-parameters used by \citet{liu2021swin}.

\paragraph{Image classification finetuning.} Hyperparameters used for finetuning each of the specific image classification datasets reported in Table~\ref{tab:classification} is provided in Appendix \ref{sec:hyperparameters}. %

\section{Conclusion}

In this paper, we have raised the question of how to pre-train models with self-supervised learning, wondering in particular on whether large scales datasets such as Imagenet are necessary for pre-training.  
Our study on ImageNet shows that taking a smaller pre-training dataset does not lead to big performance drop for denoising autoencoders, as opposed to instance discrimination self-supervised techniques or supervised pre-training.
Similarly, training on non object-centric images does not impact the downstream task performance significantly. 

Building upon these observations, we have pre-trained models directly on the target task data, instead of ImageNet, and performed evaluations on datasets of various sizes.
We have shown that it is possible to pre-train on datasets 10x smaller than ImageNet, for example obtaining +0.5 box AP gains by solely using COCO images.
\emph{We believe that this is strong evidence that large scale datasets, such as ImageNet, are not necessary for self-supervised pre-training when using denoising autoencoders.}

\paragraph{Acknowledgement.} We thank Armand Joulin, Jakob Verbeek, Natalia Neverova and Gabriel Synnaeve for fruitful discussions around this project.

\begingroup
    \setlength{\bibsep}{4pt}
    \bibliography{egbib}
    \bibliographystyle{IEEEtranN_fullname}
\endgroup

\clearpage
\twocolumn[{%
 \centering
{\Large \bf  Are Large-scale Datasets Necessary for Self-Supervised Pre-training? \\ \vspace{0.5cm} \large Appendix \par}

  \vspace*{24pt}
  {
  \par
  }
  
}]
\appendix
\pagenumbering{Roman}  

\section{\ours vs BEiT}
\label{sec:splitmask_v_beit}
We ablate our proposed components in \ours compared to a BEiT baseline in Table~\ref{tab:ablations}. All models use a ViT-B backbone and pre-trained for 300 epochs. First, we observe that the ImageNet finetuning performance improves with a margin (+0.5) by simply adopting the encoder-decoder architecture and processing two disjoint subsets per iteration. Second, the global contrastive loss on its own, without the \gls{mim} objective, provides a very weak performance. This is expected since there is no training signal for the local patch representations, and a global matching objective with 50\% masking of patches may be too hard, providing a noisy training signal and hindering the model's ability to learn informative features. 

Our full \ours model that uses both the \gls{mim} and contrastive objectives obtains the best performance and outperforms BEiT by a large margin of +0.8. The Linear probing performance of \ours is stronger than BEiT. However, both models provide a relatively weak performance on this benchmark compared to instance discrimination methods, whose final layers are more aligned to the classification task. Note, \ours adds a negligible computing overhead compared to the BEiT baseline: its wall-clock training time is marginally higher as detailed in Table~\ref{tab:ablations}. All models are trained using 16 GPUs and batch size of 2048.

\begin{table}[h]
    \centering
    \vspace{4mm}
    \caption{Ablations of different components in our \ours model in comparison with a BEiT baseline. All models including the baseline have been trained for 300 epochs using a ViT-B backbone. %
}
    \scalebox{0.85}{
    \begin{tabular}{c| c c c | c  c | c }
        \toprule
        Method & Split & Inpaint & Match  & Finetune & Lin. & Hours \\
        \midrule
        BEiT \cite{bao2021beit} & \xmarkg & \cmark & \xmarkg & 82.8 &  41.0 & 32.5 \\
        \midrule
        \multirow{3}{*}{\ours} & \cmark & \cmark & \xmarkg & 83.3 & 46.4 & \textbf{31.0} \\
        & \cmark & \xmarkg & \cmark & 79.3 & 4.0 & 32.5 \\
        \rowcolor{Goldenrod}
        & \cmark & \cmark & \cmark &  \textbf{83.6}  & \textbf{46.5 }& 34.0 \\
        \bottomrule
    \end{tabular}}
    \label{tab:ablations}
\end{table}

\section{Encoder-Decoder vs BEiT}
\label{sec:enc_dec_app}
\begin{figure}[htb]
    \centering
    \includegraphics[trim={1.0cm 0 2.0cm 0.5cm}, width=0.82\linewidth]{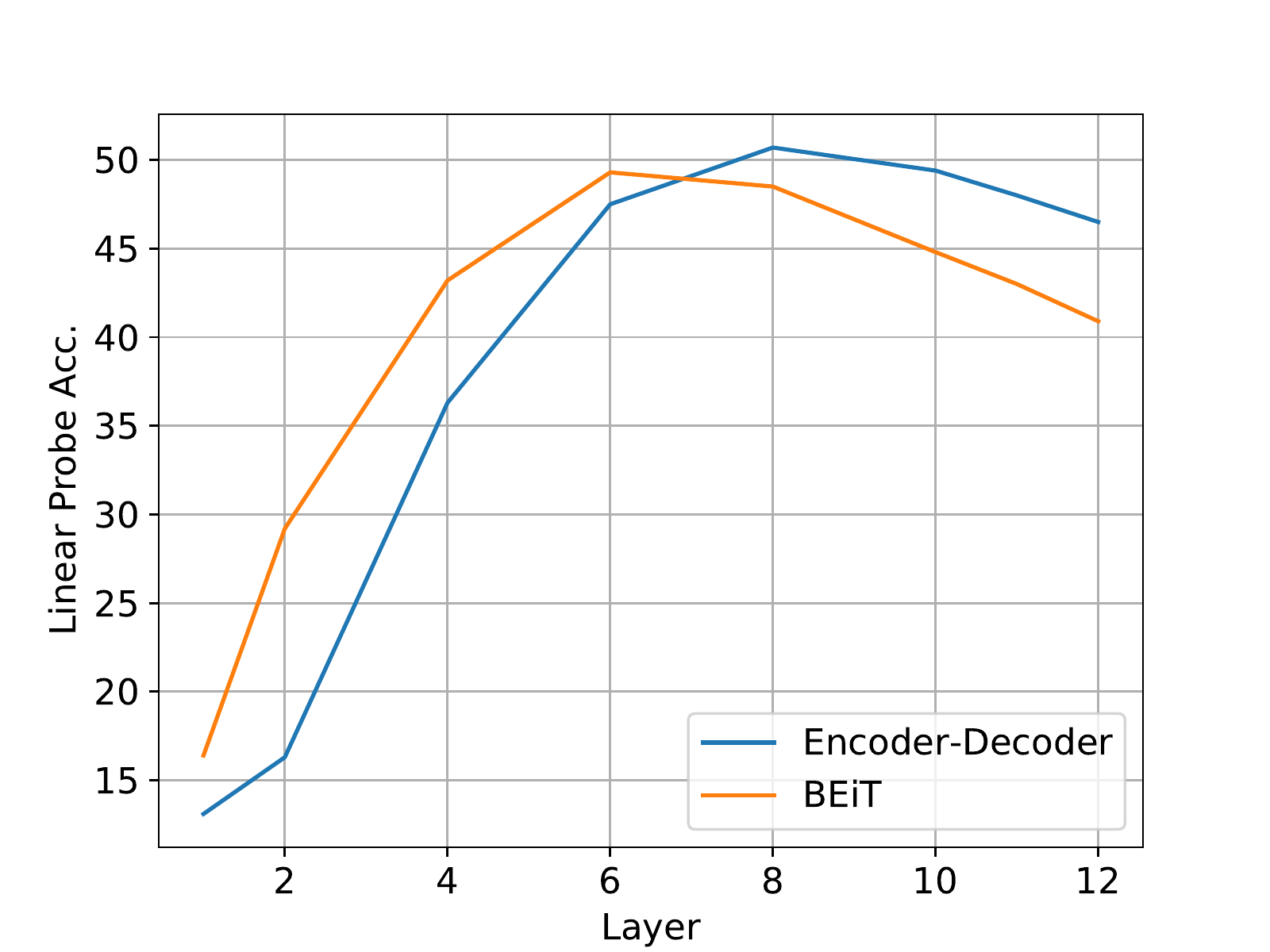}
    \caption{Linear probing accuracy on ImageNet for \ours and BEiT using features extracted from different layers.}%
    \label{fig:siam_v_beit}
\end{figure}

An advantage of the  encoder-decoder design we propose in \ref{sec:enc_dec} is that it encourages decoupling of general-purpose encoding of image features, which is required for the downstream tasks, and features specific to solving the pretext task of \gls{mim}. In particular, compared to BEiT the encoder is not capable of solving the pretext task on its own since it does not have access to the mask token. Therefore, it can only help solve the task by providing informative representation to the decoder which is the component responsible of solving the pretext task. We can see in Figure~\ref{fig:siam_v_beit} that this property improves the transferability of later layers representation to downstream tasks compared to BEiT which has a stronger drop in linear probing performance in later layers.

\section{Overfitting during pre-training}

\begin{figure}[htb]
    \centering
    \includegraphics[trim={1.0cm 0 3.0cm 0.5cm}, width=0.9\linewidth]{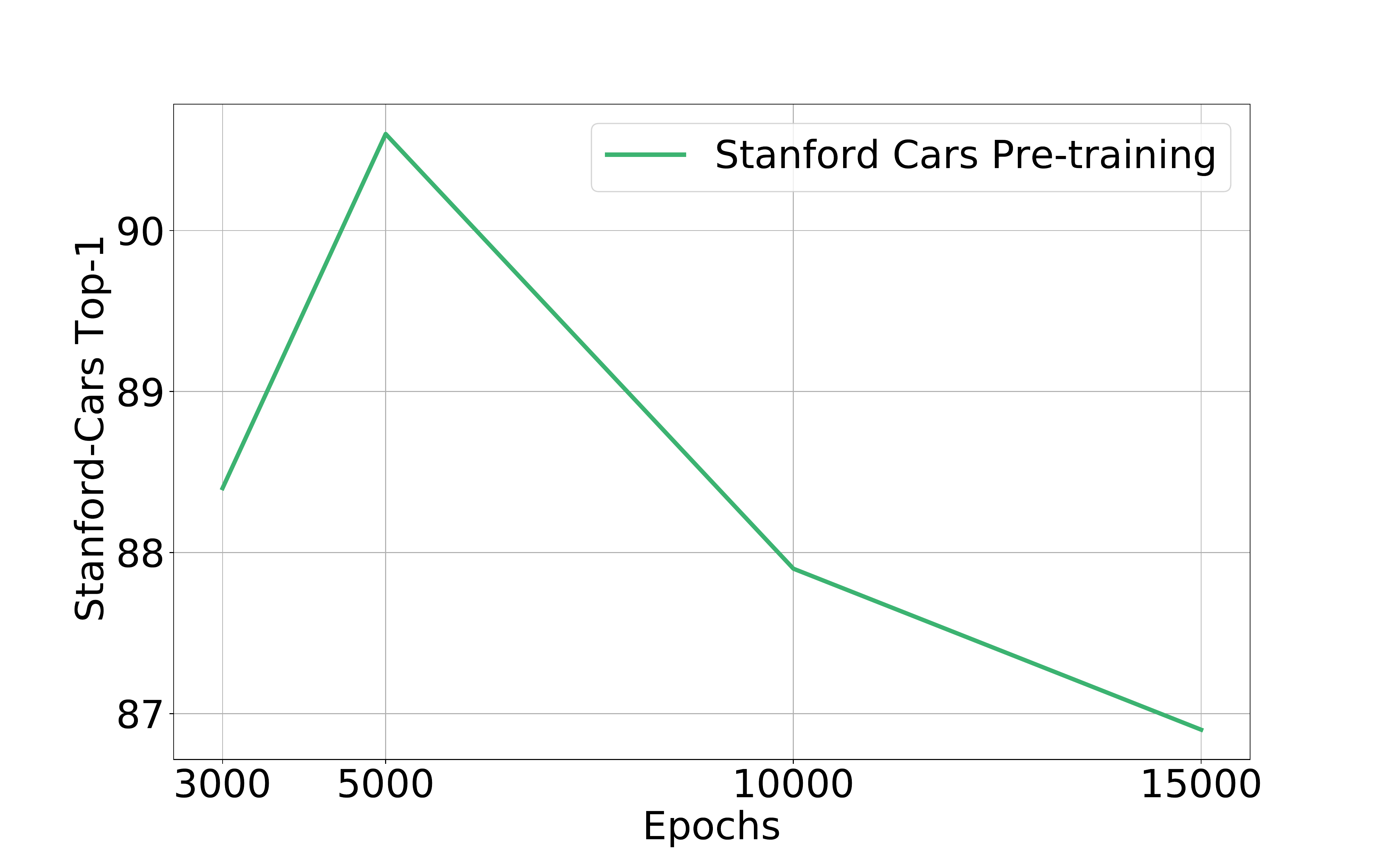}
    \caption{Finetuning performance for the Stanford Cars datasets as a function of number of pre-training epochs using the same datasets images.} %
    \vspace{4mm}
    \label{fig:cars_epochs}
\end{figure}

We observed that for pre-training of very small datasets (e.g. Stanford-Cars), longer pre-training schedules can be counterproductive. For example, if we follow the assumption we need to pre-training for the same number of updates of ImageNet pre-training for 300 epochs, the Stanford-Cars equivilant schedule would be 45k epochs. However, as we see in Figure~\ref{fig:cars_epochs}, pre-training longer than 5k epochs leads to a severe drop in finetuning performance.

\begin{table*}[ht]
\caption{
 Hyperparameters used for finetuning on the different classification datasets}
\centering
\scalebox{0.9}
{%
\begin{tabular}{l|ccc|cccc}
\toprule
Dataset & iNat18 & iNat19 & Food 101 & Cars & Clipart & Painting & Sketch \\
\midrule
Train Res & 224 & 224 & 224 & 224 & 224 & 224 & 224\\
Test Res & 224 & 224 & 224 & 224 & 224 & 224 & 224\\
\midrule
Epochs & 300 & 300 & 300 & 300 & 300 & 300 & 300\\
\midrule
Batch size & 1024 & 1024 & 1024 & 1024 & 1024 & 1024 & 1024\\
Optimizer & AdamW & AdamW & AdamW & AdamW & AdamW & AdamW & AdamW \\
Learning rate (LR) & 1.4e-4 & 1.4e-4 & 1.4e-4 & 4e-3 & 4e-3 & 4e-3 & 4e-3 \\
LR schedule & cosine & cosine & cosine & cosine & cosine & cosine & cosine\\
LR layer decay small models & \xmarkg & \xmarkg & \xmarkg & 0.65 & 0.65 & 0.65 & 0.65 \\
LR layer decay base models & \xmarkg & \xmarkg & \xmarkg & 0.65 & 0.65 & 0.65 & 0.65 \\
Weight decay & 0.05 & 0.05 & 0.05 & 0.05 & 0.05 & 0.05 & 0.05 \\
Warmup epochs & 5 & 5 & 5 & 60 & 60 & 60 & 60 \\
\midrule
Label smoothing & 0.1 & 0.1 & 0.1 & 0.1 & 0.1 & 0.1 & 0.1\\%NOLS
Dropout & \xmarkg & \xmarkg & \xmarkg & \xmarkg & \xmarkg & \xmarkg  & \xmarkg \\
Stoch. Depth & 0.1 & 0.1 & 0.1 & 0.1 & 0.1 & 0.1 & 0.1 \\
Repeated Aug & \cmark & \cmark & \cmark & \xmarkg & \xmarkg & \xmarkg & \xmarkg \\
Gradient Clip. & \xmarkg & \xmarkg & \xmarkg & \xmarkg & \xmarkg & \xmarkg & \xmarkg \\
\midrule
H. flip  & \cmark & \cmark & \cmark & \cmark & \cmark & \cmark & \cmark \\
Random Resize Crop & \cmark & \cmark & \cmark & \cmark & \cmark & \cmark & \cmark \\
Rand Augment (magnitude/std) & 7/0.5 & 7/0.5 & 7/0.5 & 9/0.5 & 9/0.5 & 9/0.5 9/0.5  & 9/0.5 \\
Auto Augment & \cmark & \cmark & \cmark & \cmark & \cmark & \cmark & \cmark \\

Mixup alpha & 0.8 & 0.8 & 0.8 & 0.8 & 0.8 & 0.8 & 0.8 \\
Cutmix alpha & 1.0 & 1.0 & 1.0 & 1.0 & 1.0 & 1.0 & 1.0 \\
ColorJitter & 0.4 & 0.4 & 0.4 & 0.4 & 0.4 & 0.4 & 0.4  \\
\midrule
Test  crop ratio & 0.875 & 0.875 & 0.875 & 0.875 & 0.875 & 0.875 & 0.875\\
\bottomrule
\end{tabular}}
\label{tab:hyperparameters}
\end{table*}

\section{Image Classification Finetuning}
\label{sec:hyperparameters}
We detail the hyperparameters used to finetune each of the classification datasets in Table~\ref{tab:hyperparameters}.

\end{document}